# A Synthetic Over-sampling method with Minority and Majority classes for imbalance problems

Hadi A. Khorshidi and Uwe Aickelin

*Abstract*—Class imbalance is a substantial challenge in classifying many real-world cases. Synthetic over-sampling methods have been effective to improve the performance of classifiers for imbalance problems. However, most synthetic over-sampling methods generate non-diverse synthetic instances within the convex hull formed by the existing minority instances as they only concentrate on the minority class and ignore the vast information provided by the majority class. They also often do not perform well for extremely imbalanced data as the fewer the minority instances, the less information to generate synthetic instances. Moreover, existing methods that generate synthetic instances using the majority class distributional information cannot perform effectively when the majority class has a multi-modal distribution. We propose a new method to generate diverse and adaptable synthetic instances using Synthetic Over-sampling with Minority and Majority classes (SOMM). SOMM generates synthetic instances diversely within the minority data space. It updates the generated instances adaptively to the neighbourhood including both classes. Thus, SOMM performs well for both binary and multiclass imbalance problems. We examine the performance of SOMM for binary and multiclass problems using benchmark data sets for different imbalance levels. The empirical results show the superiority of SOMM compared to other existing methods.

*Keywords—Imbalanced data, Synthetic over-sampling, Classification, Machine learning.*

## I. INTRODUCTION

Interesting phenomena for prediction are usually rare, high-risk or undesirable which by their nature create a minority class. Typical examples for such phenomena are software failures [1], cancer patients [2], post-surgery mortality [3], face recognition [4], image classification [5] and online fraud [6]. In these examples, the number of instances is not equally distributed between classes and at least one class has substantially different frequency. This situation is known as class imbalance and is one of the main challenges for classification algorithms [7]. Standard classifiers perform under the assumption that classes are approximately balanced. When one class has a considerably higher number of instances, the trained classification model will be biased toward that class due to a high prior probability.

H. A. Khorshidi (corresponding author) and U. Aickelin are with School of Computing and Information Systems at the University of Melbourne, Melbourne, VIC 3010 Australia (e-mails: hadi.khorshidi@unimelb.edu.au; uwe.aickelin@unimelb.edu.au).

This bias makes the classification model perform poorly in terms of distinction amongst classes, while the accuracy may remain high. The same reduction in performance happens when one class has low number of instances and a low prior probability [8].

Several methods have been developed to address the class imbalance issue given its importance in data mining. These methods can generally be grouped into two categories as classifier-design and data-balancing. Recently, generative adversarial networks (GANs) have been used to tackle imbalance problems in deep learning by considering both categories [9]. The classifier-design category aims to adjust the training process of the classifiers by considering the minority class without changing the training data set. Developing cost-sensitive classifiers is one approach in this category. Cost-sensitive classifiers assign different costs to misclassification errors of different classes as the undesirability of the misclassification varies from one class to another. For instance, classifying a patient whose risk of mortality after surgery is high as a low-risk patient is more critical than the other way around. So, the classifier minimises the cost-based weighted misclassification error instead of minimising the regular misclassification error [10], [11]. Another approach is thresholding that alters the decision boundary in classification [12].

The data-balancing category includes methods that re-samples the training data set to balance the classes [13]. Random under-sampling and random over-sampling are the most basic methods for re-sampling. The main shortcomings of these methods are the loss of valuable information and increasing the chance of overfitting, respectively [14], [15]. Another shortcoming of random over-sampling is adding a number of identical instances into the training data which is a challenge for classifiers [7]. Random over-sampling can be performed in a focused way, i.e., minority instances that are close to the boundary between the classes are replicated. However, the problem of adding identical instances still holds. In addition, focused over-sampling results in the construction of smaller and more specific decision regions [14].

To overcome these shortcomings, synthetic over-sampling methods have been introduced. These methods have been effective to improve the performance of classifiers when class imbalance exists [15], [16]. The Synthetic Minority Over-sampling Technique (SMOTE) [17] is the most commonly used methods and generates synthetic instances using nearest neighbours of the minority class. So, the synthetic instances broaden the data space for minority instances. SMOTE has shown success in stabilising the performance of classifiers [18] and has been extended to develop a suite of algorithms known as the SMOTE family [19]. However, as we will explain in the next section, limitations remain for SMOTE family algorithms and other synthetic over-sampling methods, which motivates our study.

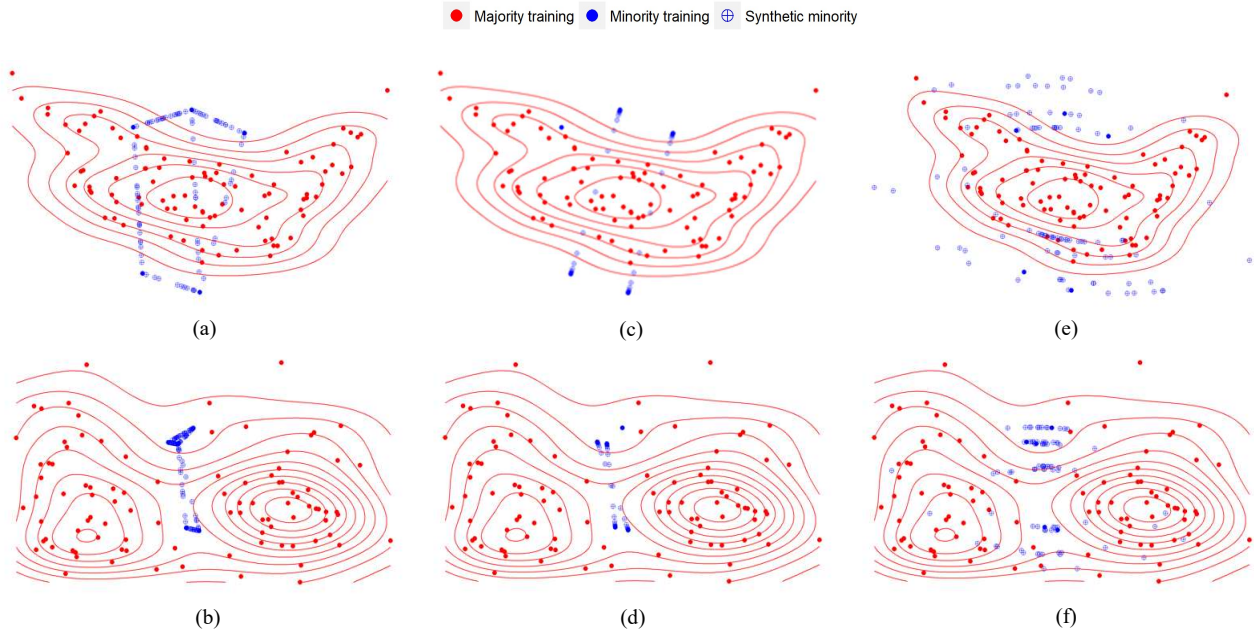

Fig. 1. Two examples for illustration of synthetic generation using SMOTE (a & b), MAHAKIL (c & d) and SWIM (e & f).

*A. Motivation*

In this section, we review the limitations of the existing synthetic over-sampling methods. We use two examples (shown in Fig. 1) to illustrate these limitations. Example 1 visualises the situation that the data space for the majority class is a convex hull, while the minority class consists of two separate regions. Example 2 visualises the situation that the data space of the majority class is bi-modal, and the region of the minority class is a convex hull located between two high-density regions of the majority class.

One of the limitations of SMOTE is that it generates synthetic instances within the convex hull formed by the minority instances existing in the training data set. This can lead to generating synthetic instances in the regions belonging to the majority class (over-generalisation). Fig. 1. (a) shows how SMOTE generates instances in the data space of the majority class, especially in the region with higher density that should be avoided to prevent reduc-ing the performance of classifiers.

Many variants have been developed for SMOTE such as Borderline-SMOTE [20], Adaptive Synthetic Sampling (ADASYN) [21], density-based SMOTE (DBSMOTE) [22], Evolutionary Cluster-Based Synthetic Oversampling Ensemble (ECO-Ensemble) [23] and Majority Weighted Minority Over-sampling Technique (MWMOTE) [24] to improve its performance. SMOTE and its variants use similar procedures to generate synthetic instances using nearest neighbours. Generating synthetic instances based on nearest neighbours results in having less diverse synthetic data [18] and complications in covering the data space of the minority class [25]. Keeping diversity in re-sampling methods improves the performance of classifiers for imbalanced data [26], [27]. Fig. 1 (a) and (b) show this limitation of SMOTE as the synthetic in-stances are linear interpolations of existing minority in-stances.

MAHAKIL [18] is a synthetic over-sampling method that has been introduced recently to overcome the limitations of the SMOTE family notably by generating more diverse synthetic instances. MAHAKIL generates synthetic instances by pairing existing minority instances and previously generated synthetic instances as parents to generate new instances inspired by the Chromosomal Theory of Inheritance [28]. Fig. 1 (c) and (d) show the performance of MAHAKIL to generate synthetic instances for both examples. Fig. 1 (c) shows that MAHAKIL performs better than SMOTE by generating less instances in the data space of the majority class. However, both examples show that MAHAKIL cannot generate diverse synthetic instances when the number of instances in the minority class is low. This is because MAHAKIL can only assure the diversity when there are enough and diverse parents to select from.

Another limitation of both SMOTE and MAHAKIL is that they only use the existing minority instances to gen-erate synthetic instances. Thus, they ignore the vast in-formation provided by the majority class. Recently, synthetic over-sampling with the majority class (SWIM) [15] has been introduced to capture majority class information for synthetic instance generation. SWIM uses Mahalanobis distance [29] to generate synthetic instances that are in the same distance from the mean of the distribution of the majority class as the existing minority instances. Fig. 1. (e) shows SWIM can successfully generate synthetic instances that avoid the higher density region of the ma-jority class even though some of the synthetic instances are in its data space. However, SWIM cannot ensure avoiding the generation of synthetic instances in the higher density region of the majority class when its dis-tribution is bi-modal (Fig. 1 (f)). This is because the mean of the distribution is not located in the higher density region of the bi-modal distribution. In addition, the syn-thetic instances generated by SWIM are not diversely distributed in the data space of the minority class.

On a related note, most over-sampling methods, includ-ing the SMOTE family and MAHAKIL, do not perform well in cases of extreme imbalance, because they rely on minority instances to generate synthetic instances. Thus, the fewer the minority instances, the less information is provided to generate synthetic instances. SWIM is a powerful method to generate synthetic instances when the minority instances in the training data set are limited. However, its performance deteriorates in comparison with existing synthetic over-sampling methods as the number of minority instances increases [15].

Finally, a general concern about synthetic over-sampling methods is to widen the data space of minority class by generating synthetic instances in the regions of the majority class. This would not be crucial when the number of these synthetic instances is few. However, if the generated instances either are in higher density regions of the majority class or form a new decision region, they can provide misleading information for classifiers.

### B. Contributions

To address the limitations and concerns mentioned in the motivation section, we define our research questions (RQs) as follows:

RQ1: How to generate synthetic instances adaptable to the data space of the majority class for cases such as non-convex minority data space?

RQ2: How to generate synthetic instances diversely to cover the data space of the minority class?

RQ3: Given the data space of both minority and majority classes, where should the synthetic instances be generated?

RQ4: Is it possible to have a synthetic over-sampling method that can perform appropriately for high imbalance (around 5%), extreme imbalance (1% and less) and absolute imbalance (few minority instances) data?

In this study, we aim to propose a new method to generate synthetic instances to address the defined research questions. The proposed method is Synthetic Over-sampling with the Minority and Majority classes (SOMM). SOMM does not use the nearest neighbours of the minority class to generate synthetic instances as the SMOTE family does. Thus, SOMM can generate diverse synthetic instances within the data space of the minority class. However, SOMM considers the nearest neighbours from both minority and majority classes when it locates the synthetic instances by maintaining the diversity. As a result, SOMM remains adaptable to the data space of the majority class and non-convexity of the data space of the minority class. These properties provide a capability for SOMM to improve classifiers for multiclass imbalance problems as well.

---

**Algorithm 1: SOMM algorithm**

Input: $D$ training data, $N_S$ number of synthetic instances, $k$ number of nearest neighbours of a generated instance
Output: A set of synthetic instances $S_u$

1. Normalise the training data ($D$) using (1) as $D_{norm}$
2. Separate majority and minority subsets as $A_{norm}$ and $B_{norm}$ respectively
3. Determine the vectors of upper and lower bounds for features in $B_{norm}$ as $u_f$ and $l_f$ respectively
4. $S_u, S$ = Null; $i = 1$
5. While $i <= NS$ do
6.    for f features do
7.       $S_{if}$ ← Generate a random value between uf and lf
8.    end for
9.    $nn$ ← Find k nearest neighbours of $S_{ij}$ in $D_{norm}$
10.   if $nn \cap B_{norm} = \emptyset$ then $i \leftarrow i-1$ /* Remove the generated instance */
11.   else $nn^{sort}$ ← Sort the instances in $nn$ based on their distance from $S_{ij}$
12.   if $nn_1^{sort} \epsilon\ B_{norm}$ then $S_u \leftarrow S_{ij}$ else
13.     $index_B$ ← Find the index for the closest minority neighbour in $nn^{sort}$
14.     $nn^{sort}$ ← Keep the first indexB instances in $nn^{sort}$
15.     $dis_B$ ← Calculate the distance between $nn^{sort}_{index_B}$ and $S_{ij}$
16.     $dir$ ← $nn^{sort}_{index_B} - S_{ij}$
17.     for j 1 to $index_B$ -1 do
18.       $dis_P$ ← Calculate the projected distance of the majority neighbours from Sij using (2) and (3)
19.     end for
20.     $Max_{disp}$ ← max $dis_P$
21.     $m$ ← Generate a random value between $Max_{disp}$ and $dis_B$
22.     $S_u$ ← Update $S_{ij}$ using (4)
23.     $i \leftarrow i+1$
24.   end while
25. De-normalise $S_u$ using (5)

---

SOMM can generate diverse synthetic instances for each minority class that are accountable to other classes. The main attention in learning from imbalanced data has been on binary problems, while dealing with multiclass problems are relatively more challanging [30]. Normally, the methods that work for binary problems do not perform properly for multiclass problems. However, in this paper, we show that SOMM can perform effective for both binary and multiclass problems.

We empirically examine the ability of SOMM to generate diverse synthetic instances to cover the minority space using 6 synthetic data sets. We also validate the performance of SOMM in comparison with existing recent methods using 26 benchmark data sets, including 6 synthetic and 20 real data sets, in high, extreme and absolute situations for binary imbalance problems and 10 data sets for multiclass imbalance problems. To encourage further take-up, we also ensure that SOMM is formulated intuitive and relatively simply and can be applied to different types of data.

## II. METHOD

In this section, we describe the details of our SOMM algorithm (Algorithm 1[1]) for binary imbalance problems. This algorithm can be expanded for multiclass imbalance problems by targeting one of the minority classes as the minority class and all other classes as the majority class. Then, we discuss the advantages of SOMM in comparison with other synthetic over-sampling methods.

### A. Algorithm

The algorithm's aim is to generate synthetic instances via addressing RQs 1-3. The generated synthetic instances diversely cover the data space of the minority class. In addition, the algorithm is adaptable to the neighbourhood. This algorithm adapts to the neighbourhood by takin into account both minority and majority data spaces when it confirms the placement of the generated synthetic instance. Consequently, SOMM would perform well when the number of minority instances are enormously low in comparison with majority instances. We describe the five steps of SOMM in details as follows. The training data set is indicated as D, the number of synthetic instances that need to be created to balance the training data set is denoted by $N_S$.

#### 1) Normalise the data set

Before generating the synthetic instances, we transform the data set so that all features are ranged between zero and one. This transformation enables the application of Euclidean distance (as being widely used in the literature) in the fourth step to update the generated synthetic instances properly. The z-score normalization does not work as SOMM does not consider distributional information (this has been investigated empirically as well). Let $Max_D$ and $Min_D$ be the vectors of maximum and minimum values for features, respectively. Thus, the normalised data set ($D_{norm}$) is obtained using (1).

$$D_{norm} = (D - Min_D) \div (Max_D - Min_D) \qquad (1)$$

#### 2) Set feature bounds

After the transformation in the first step, we separate $D_{norm}$ into two subsets of majority and minority classes. Let $A_{norm}$ and $B_{norm}$ be the transformed subsets of majority and minority, respectively. Then, for each feature in $B_{norm}$, we set maximum and minimum values as upper ($u_f$) and lower ($l_f$) bounds.

#### 3) Generate synthetic instances

We randomly generate a value between $l_f$ and $u_f$ for each feature. Thus, we generate synthetic instances ($s$) as vectors where all elements lie within feature bounds. Consequently, the generated instances can be diversely spread across the data space of the minority class (addressing RQ2).

#### 4) Update the generated instances

Although we generate the instances inside the feature bounds, they may be placed outside the data space of the minority class. For instance, if the minority class is not convex, the space bounded by the feature bounds may overlap the data space of the majority class. So, some of the synthetic instances that are generated in step 3 may breach the border and end up in the regions of the majority class.

We address this issue by investigating the neighbourhood of the generated instance. Once an instance ($s$) is generated, we find $k$ nearest neighbours of that generated instance. These neighbours could be from both majority and minority classes. We apply the three following rules to prevent the widening of the data space of the minority class by generating synthetic instances in the regions of the majority class.

1) If all the neighbours belong to the majority class, we remove that generated instance (addressing RQ1).
2) If the condition of rule (1) is not met, we order the neighbours based on their distance from the generated instance. If the closest neighbour belongs to the minority class, we keep that generated instance and update it as $s_u$ (updated instance).
3) If the condition of rule (2) is not met, we cut the ordered set of neighbours at the point of the closest neighbour from the minority class. After that, we would have a subset ($nn$) of the ordered set that includes a set of the closest neighbours from the majority class ($nn_A$) and ending with the closest neighbour from the minority class ($nn_B$).

We assimilate the generated instance to the minority class by moving from the neighbourhood of the majority class toward $nn_B$. To do this, we need to define the direction and the magnitude of the movement. The direction ($dir$) is the vector from the generated instance to $nn_B$. If the magnitude ($m$) equalises with the distance between $nn_B$ and the generated instance ($dis_B$), $nn_B$ would be replicated. As we want to maintain the diversity in the over-sampled data, we avoid replicating $nn_B$. Thus, we set the magnitude in a way that the generated instance moves away from the neighbourhood of the majority class (addressing RQ3). To do this, we need to project the neighbours in $nn_A$ onto $dir$. First, we calculate the Cosine of the angle ($\theta$) between $dir$ and the vectors from the generated instance to the neighbours in $nn_A$ ($dir_A$) using (2). Then, we find the projected distance ($dis_P$) of each neighbour in $nn_A$ onto $dir$ using the distance of each neighbour from the generated instance ($dis_A$) as (3).

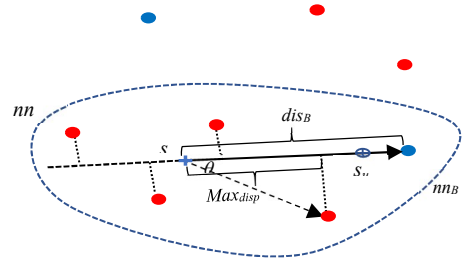

$$Cosine\ \theta = (dir_A \cdot dir) \div (|dir_A|_2 \times |dir|_2) \qquad (2)$$
$$dis_P = dis_A \times Cosine\ \theta \qquad (3)$$

where $||_2$ indicates the norm 2 that calculates the size of each vector. Also, Cosine $\theta$ and $dis_P$ are sets with the length of $nn_A$.

---
[1] The codes are provided at https://github.com/hadi1453/Synthetic-Over-sampling-with-the-Minority-and-Majority-classes-for-imbalance-problems/blob/main/SOMM.R

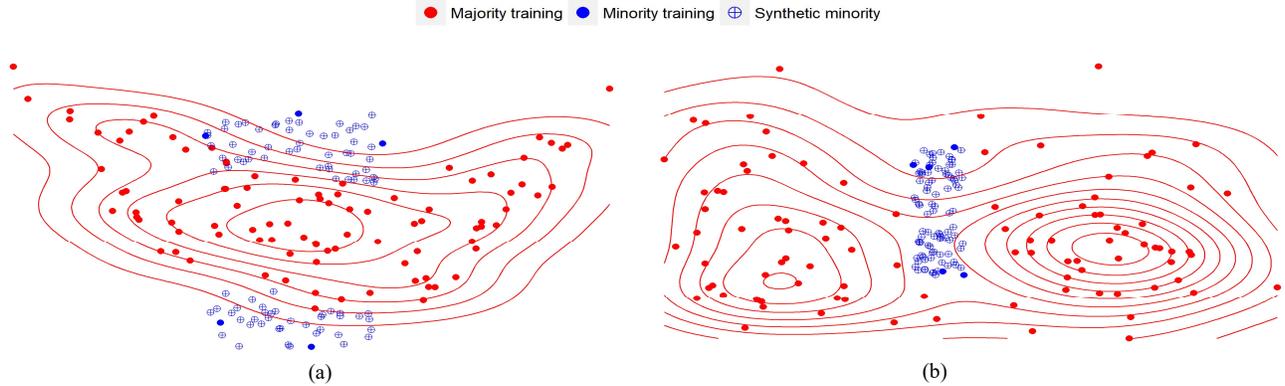

Fig. 3. Two examples for illustration of synthetic generation using SOMM.

We obtain the maximum value in $dis_P$ ($Max_{disp}$). Then, we randomly select a number between $Max_{disp}$ and $dis_B$ to set as the magnitude. Lastly, the generated instance is updated using (4).

$$s_u = s + dir \times m \quad (4)$$

Fig. 2. An illustrative example for updating the generated instances; red and blue circles are instances from majority and minority classes respectively, + denotes the generated instance and ⊕ is the updated generated instance.

Fig. 2 shows the process of updating the generated instance described in step 4 of the SOMM algorithm. The generated instance + is updated as ⊕ by moving towards the minority instance (blue circle) with the extent of magnitude between $Max_{disp}$ and $dis_B$.

*5) Transform back to the original space*

As the final generated instances ($s_u$) are still in the normalised space, we need to transform them into the original space using (5). Hence, they can be added to the training data set for constructing classifiers.

$$s_{SOMM} = s_u \times (Max_D - Min_D) + Min_D \quad (5)$$

where $s_{SOMM}$ are the synthetic instances generated by SOMM method for over-sampling the minority class.

*B. Comparative discussion for binary imbalance problems*

In this section, we compare SOMM with SWIM, MAHAKIL and the SMOTE family in terms of generating synthetic instances. We choose these methods as they are synthetic over-sampling methods and appropriate benchmarks to examine different properties of SOMM. SMOTE is the most popular synthetic over-sampling method. MAHAKIL considers diversity and SWIM incorporates distributional information from majority class in generating synthetic minority instances. Ideally to construct a robust classifier, the generated synthetic instances should correspond with the real data space of the minority class. We also apply SOMM on two examples used in Fig. 1. Fig. 3 shows how SOMM performs on these examples. In Fig. 3 (a), we can see that SOMM keeps the non-convex space of the minority class. In both figures, there is no synthetic instances generated in the higher density regions of the majority class. Moreover, SOMM can generate diverse synthetic instances to cover the data space of the minority class even when the number of existing instances is low.

*1) SOMM vs SWIM*

The main difference between these two methods is that SWIM relies on distributional information of majority class' space and its mean value to generate synthetic instances. So, SWIM's performance deteriorates when the majority class is not a convex hull (Fig. 1 (f)). However, SOMM can manage generating synthetic instances that are not in the higher density regions of the majority class even if there is a bi-modal space for the majority class (Fig. 3 (b)).

Another difference is that SOMM generates synthetic instances diversely across the data space of the minority class. On the other hand, SWIM only generates synthetic instances in the same hyper-elliptical density contour around the majority class as the minority instances.

SWIM's algorithm involves the computation of matrix inverses and square roots of those inverses. This leads to failing the computation if the features are linearly dependent. To overcome this issue, there is a need to identify and remove those features, and proceed with SWIM's algorithm on the sub-space with independent features [15]. Thus, a pre-processing step is needed and there would be an information loss. On the contrary, SOMM can be applied more simply on various types of data sets.

*2) SOMM vs MAHAKIL*

The main difference between these two methods is that MAHAKIL does involve information from the majority class. In addition, it just generates synthetic instances within the convex hull formed by the existing minority instances. Thus, MAHAKIL may widen the data space of the minority class mistakenly into the majority space. However, SOMM considers the information of the neighbourhood from both majority and minority classes to avoid generating instances within the majority space.

Even though MAHAKIL was originally proposed to generate diverse synthetic instances inside the data space of the minority class, it fails to generate diverse instance when there are few minority instances (Fig. 1). The MAHAKIL's algorithm does not work if the number of minority instances is less than four. Whereas even in these cases, SOMM can cover the minority space diversely by generating synthetic instances.

*3) SOMM vs SMOTE*

The main difference between SOMM and the SMOTE family is that SMOTE and its variants generate synthetic instances using the nearest neighbours of the minority class. Although some of the methods in the SMOTE family incorporate the information of the majority class in cleaning the generated instances, they still rely on generating synthetic instances through SMOTE [15]. Thus, the limitations of generating non-diverse synthetic instances within the convex hull of the minority instances are still hold.

In addition, there are some specific limitations with each method in the SMOTE family. Borderline-SMOTE [20] uses the information of the majority space to generate synthetic instances on the borderline between two classes. This leads to creating higher density regions for the minority class on the borderline that may not correspond with the real data space of the minority class and be misleading for classifiers. ADASYN [21] and MWMOTE [24] extend Borderline-SMOTE by giving weights to minority instances where on or nearby the borderline. ADASYN calculates a density distribution and MWMOTE defines specific weights to generate more synthetic instances for minority instances in the neighbourhood of the majority class. Both create regions of higher density for minority class around the borderline. MWMOTE also generates synthetic instances via clustering the minority instances. This results in deterioration of MWMOTE's performance when the number of minority instances is limited as there is not enough information for constructing the clusters. MWMOTE has six parameters that should be pre-defined, while SOMM only needs one pre-defined parameter.

*C. Comparative discussion for multiclass imbalance problems*

In this section, we introduce two comparable synthetic over-sampling methods for multiclass imbalance problems. These methods are Mahalanobis Distance-based Over-sampling technique (MDO) [31] and k-nearest neighbours (k-NN)-based synthetic minority oversampling algorithm (SMOM) [32]. Then, we discuss the advantages of SOMM over MDO and SMOM.

MDO generates synthetic instances for each minority class using Mahalanobis distance and distributional information of that minority class within the ellipse contour that passes through the chosen instances of the minority class. To choose these instances, for every instance, $K1$ nearest neighbours are obtained. If there are more $K2$ neighbours that belong to the minority class, the instance is chosen, and a weight is assigned to that instance. $K1$ and $K2$ are predefined parameters of MDO. Then, the chosen instances are scaled to have zero mean and transformed to principal components (PCs) spaces. The synthetic instances are generated randomly on the ellipse contours in the PC spaces. Finally, the generated instances are transformed back into the original space.

SMOM generates synthetic instances for each minority class using $k$-NN of the minority instances like SMOTE through linear interpolation. However, to avoid over-generalisation, SMOM clusters the minority instances using a neighbourhood-based clustering algorithm (NBDOS) which is derived from a density-based clustering method, DBSCAN [33]. NBDOS groups minority instances based on the density of minority instances in their neighbourhood. The instances that are in the regions that are dominated by the minority class are outstanding instances and others are trapped instances. Then, weights are assigned to trapped instances based on rules defined based on the neighbourhood.

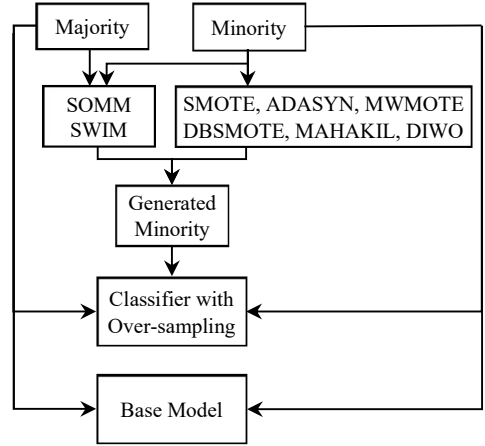

Fig. 4. The framework of experimental design.

Both MDO and SMOM use computationally intensive algorithms. MDO examines the neighbourhood of every single instance in all minority classes to choose appropriate instances and assign weights. SMOM also examines the neighbourhood of all instances in minority classes to find their cluster. Then, all trapped instances are examined by new rules to assign weights to be involved in generating synthetic instances. SOMM, on the other hand, is formulated relatively simply.

SMOM needs eight pre-defined parameters such as $k1$: the number of nearest neighbours to generate the synthetic instances, *k2*: the number of nearest neighbours for clustering, *rTh*: the minimal proportion of the minority class instances in their *k2*-nearest neighbours to be assigned as a centre of cluster, *nTh*: the minimal number of members to form a cluster, *w1*, *w2*, *r1*, and *r2*: the parameters used for calculating the weights. MDO needs two parameters, and SOMM just needs one pre-defined parameter.

TABLE I
DATA SETS FOR BINARY IMBALANCE PROBLEMS

| Dataset | Name | Dimension | Size | Ratio[*] | Dataset | Name | Dimension | Size | Ratio |
|---|---|---|---|---|---|---|---|---|---|
| D1 | Wisconsin | 9 | 683 | 65-35 | D11 | Vehicle Bus | 18 | 846 | 74-26 |
| D2 | Waveform 0 | 21 | 1,800 | 67-33 | D12 | Pima | 8 | 768 | 65-35 |
| D3 | Vowel 0 | 10 | 990 | 91-09 | D13 | Wine White High[§] Quality | 11 | 4,898 | 78-22 |
| D4 | Vowel 1 | 10 | 990 | 91-09 | D14 | Wine White Low vs High | 11 | 1,243 | 85-15 |
| D5 | Vowel 2 | 10 | 990 | 91-09 | D15 | Wine Red Low[§] Quality | 11 | 1,599 | 96-04 |
| D6 | Vowel 10 | 10 | 990 | 91-09 | D16 | Wine Red Low vs High | 11 | 280 | 77-23 |
| D7 | Poker 7-9 vs 6 | 10 | 1,669 | 85-15 | D17 | Thoracic Surgery | 24 | 470 | 85-15 |
| D8 | KDD Control | 60 | 600 | 83-17 | D18 | Norm Ring | 20 | 5,532 | 66-34 |
| D9 | Oil | 49 | 937 | 95-05 | D19 | Spambase | 57 | 4,601 | 61-39 |
| D10 | Forest Cover 3 | 54 | 3,000 | 95-05 | D20 | Abalone 9-18 | 9 | 2,676 | 68-32 |

[*]Ratio denotes the percentages of majority-minority instances in the original data set.
[§]High refers to the quality values of 7 and above; Low refers to the quality values of 4 and less.

methods using the smotefamily R package [34] and MWMOTE using the imbalance R package [35]. We keep the parameters as default. For SWIM, DIWO and MAHAKIL, we develop the algorithm as described by the authors in the papers [15], [18], [27], and set the parameters as which is reported as the best values.

*1) Diversity*

In this section, we want to examine the performance of MDO does not have an effective mechanism to avoid generating synthetic instances in the regions belonging to other classes, especially when the chosen instances are selected based on Euclidean distance and synthetic instances are generated in a non-Euclidean space. In contrast, both SOMM and SMOM have mechanisms to avoid generating instance on other classes' region.

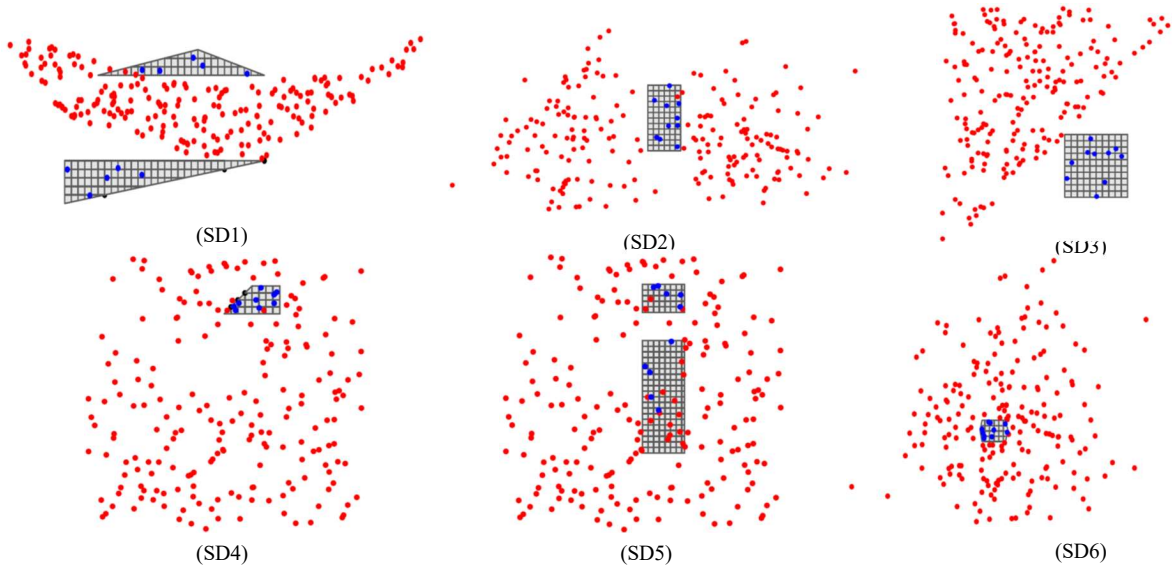

Fig. 5. Synthetic data sets.

In the following sections, we describe the experiments, present the results and discuss the findings for both binary and multiclass imbalance problems.

## III. EXPERIMENT RESULTS FOR BINARY IMBALANCE PROBLEMS

### A. Experimental design

We design experiments to compare the performance of SOMM with other synthetic over-sampling methods for binary imbalance problems. We choose SMOTE as the most commonly used synthetic over-sampling method. We also choose ADASYN, MWMOTE and DBSMOTE as methods in SMOTE family, MAHAKIL and Diversity-based Instance-Wise Over-sampling (DIWO) [27] as two recently introduced methods which focus on diversity, and SWIM which uses majority distribution to generate synthetic instances. Fig. 4 shows the framework of experimental design using minority and majority information via synthetic over-sampling methods to construct classifiers.

We implement SMOTE, ADASYN and DBSMOTE

SOMM in generating diverse synthetic instances within the minority data space (examining RQ2). In previous works, this property has been discussed using visualisation [15, 25] as the data space of the minority class is not normally accessible. However, we design an experiment using synthetic data sets (SDs) to measure how well each over-sampling method can cover the minority space. We create 6 data sets that represent different situations such as non-convex minority space, bi-modal majority, linearly separable classes, and overlapped classes (examining RQ1).

Fig. 5 shows these data sets. In SD1, the majority space is a convex hull while the minority space is non-convex and divided by the majority space. Two classes have small overlapping. In SD2, the majority space is bi-modal while the minority space is located between high-density regions of the majority class as a convex hull. The minority space does not overlap with the high-density regions of the majority class. In SD3, both minority and majority spaces are convex and separated linearly from each other. In SD4, the minority instances are inside the majority space and both spaces are convex. In SD5, the minority space is non-convex and is within the majority space. In SD6, both spaces are convex and the minority space is within the majority space and overlaps with high-density region of the majority instances.

To measure the diversity of synthetic instances within the minority space, inspired by spatial coverage measures [36], we divide the minority space into cells. Then, we count the number of cells that do not contain any synthetic instances (not covered cells, NCC). We measure the coverage diversity (CovDiv) using

TABLE II
THE CovDiv RESULTS FOR OVER-SAMPLING METHODS

| Dataset | # | SOMM | SMOTE | MAHAKIL | SWIM | ADASYN | DBSMOTE | MWMOTE | DIWO |
|---|---|---|---|---|---|---|---|---|---|
| SD1 | 6 | **0.361±0.087** | 0.289±0.056 | 0.157±0.035 | 0.339±0.057 | 0.252±0.054 | 0.175±0.046 | 0.191±0.057 | 0.275±0.062 |
| SD1 | 10 | *0.438±0.069* | 0.356±0.052 | 0.252±0.045 | 0.374±0.062 | 0.318±0.049 | 0.232±0.048 | 0.250±0.049 | 0.344±0.059 |
| SD1 | 20 | *0.479±0.063* | 0.443±0.044 | 0.359±0.041 | 0.383±0.063 | 0.361±0.052 | 0.286±0.044 | 0.331±0.042 | 0.389±0.059 |
| SD2 | 6 | *0.707±0.162* | 0.564±0.126 | 0.311±0.07 | 0.591±0.121 | 0.559±0.128 | 0.360±0.092 | 0.463±0.158 | 0.571±0.134 |
| SD2 | 10 | *0.857±0.101* | 0.661±0.11 | 0.499±0.06 | 0.687±0.085 | 0.683±0.112 | 0.455±0.079 | 0.555±0.114 | 0.683±0.101 |
| SD2 | 20 | *0.916±0.064* | 0.759±0.087 | 0.713±0.082 | 0.707±0.068 | 0.537±0.125 | 0.525±0.082 | 0.671±0.092 | 0.791±0.078 |
| SD3 | 6 | **0.717±0.17** | 0.557±0.117 | 0.309±0.062 | 0.703±0.11 | 0.567±0.125 | 0.373±0.093 | 0.448±0.142 | 0.545±0.125 |
| SD3 | 10 | *0.856±0.098* | 0.656±0.097 | 0.485±0.07 | 0.744±0.089 | 0.684±0.109 | 0.455±0.083 | 0.510±0.154 | 0.684±0.110 |
| SD3 | 20 | *0.923±0.053* | 0.729±0.096 | 0.720±0.079 | 0.799±0.07 | 0.715±0.095 | 0.546±0.08 | 0.514±0.103 | 0.775±0.076 |
| SD4 | 6 | 0.605±0.173 | 0.493±0.113 | 0.276±0.05 | **0.629±0.136** | 0.452±0.116 | 0.324±0.096 | 0.395±0.113 | 0.491±0.118 |
| SD4 | 10 | 0.722±0.141 | 0.540±0.086 | 0.415±0.074 | 0.721±0.105 | 0.502±0.114 | 0.378±0.073 | 0.437±0.089 | 0.550±0.089 |
| SD4 | 20 | 0.826±0.089 | 0.650±0.061 | 0.633±0.063 | 0.806±0.072 | 0.441±0.092 | 0.474±0.072 | 0.577±0.073 | 0.686±0.071 |
| SD5 | 6 | *0.537±0.097* | 0.442±0.084 | 0.193±0.037 | 0.424±0.071 | 0.437±0.094 | 0.263±0.062 | 0.269±0.107 | 0.412±0.093 |
| SD5 | 10 | *0.599±0.082* | 0.499±0.073 | 0.328±0.044 | 0.496±0.064 | 0.464±0.071 | 0.299±0.061 | 0.400±0.067 | 0.512±0.065 |
| SD5 | 20 | *0.608±0.057* | 0.569±0.067 | 0.545±0.07 | 0.496±0.053 | 0.449±0.073 | 0.381±0.063 | 0.444±0.065 | 0.549±0.057 |
| SD6 | 6 | 0.635±0.177 | 0.479±0.125 | 0.269±0.056 | **0.655±0.1** | 0.483±0.127 | 0.374±0.077 | 0.397±0.108 | 0.503±0.142 |
| SD6 | 10 | *0.812±0.083* | 0.608±0.085 | 0.476±0.089 | 0.741±0.085 | 0.593±0.125 | 0.423±0.072 | 0.516±0.09 | 0.651±0.11 |
| SD6 | 20 | *0.931±0.059* | 0.745±0.079 | 0.695±0.08 | 0.840±0.071 | 0.564±0.13 | 0.533±0.066 | 0.656±0.082 | 0.789±0.066 |

(6) by calculating the ratio of cells contain synthetic instances.

$$CovDiv = 1 - \frac{NCC}{number\ of\ cells} \quad (6)$$

The CovDiv not only measures the diversity, but also ignores the synthetic instances that located outside the minority space.

For this experiment, we keep the number of majority instances as 100, but set the number of minority instances at three different values as 6, 10 and 20 to examine the methods under different situations. We choose the minimum number of minority instances as 6 because SMOTE and MWMOTE cannot work especially when K=5 as the optimal number of neighbours [15]. Also, MAHAKIL cannot work if the minority instances are less than 4. We repeat each experiment 30 times to check different possibilities of minority instances within their space for an unbiased evaluation. We also conduct Mann-Whitney test, a non-parametric statistical test, to examine the significance of the differences across over-sampling methods. We use Mann-Whitney tests as we cannot ensure the normality of g-mean values.

*2) Classification*

In this section, we examine the performance of over-sampling methods in classification task. We use 6 synthetic data sets and 20 real benchmark data sets[2] (listed in Table I) in this experimental design. We select these data sets as they can cover various ranges of size and dimension to provide a fair base of comparison. In addition, these data sets have enough minority instances to create training data sets with different levels of class imbalance as required for RQ4. For SDs, we set the number of majority instances as 1000, and minority instances as 100 to have enough instances for experiments in different levels. We create three different training data sets with 5% (high), 1% (extreme) imbalance ratios and only 6 minority instances (absolute imbalance).

We randomly divide the data sets into training and testing subsets by ratios of 75% and 25% respectively. We choose the ratio of 75% for training subset to have more minority instances in the training data. We set the minimum number of minority instances as 8 for data with 1% imbalance ratio to differentiate between extreme and absolute imbalance levels. Then, we randomly down-sample the minority instances in the training subset to reach the desired imbalance levels. This process is repeated 30 times to cover different combinations in our performance evaluation and provide a possibility for conducting statistical tests.

Once the training data are created, the synthetic over-sampling methods are implemented to balance the majority and minority classes. Then, the final data set is used to construct classifiers. We also construct classifiers without over-sampling as base models (As shown in Fig. 4). We use seven classifiers to provide a comprehensive investigation. These classifiers are Logistic Regression (GLM), Naïve Bayes (NB), Decision Trees (CART), $k$-nearest neighbour ($k$-NN), Neural Networks (NN) with single hidden layer, Random Forest (RF) and Support Vector Machine (SVM) with radial kernel. We set up all the classifiers with their default settings.

---
[2] https://archive.ics.uci.edu/ml/datasets.php;
https://web.stanford.edu/~hastie/ElemStatLearn/datasets/;
https://machinelearningmastery.com/imbalanced-classification-model-to-detect-oil-spills/

The constructed classifiers are applied on the test data sets to examine their performance. We use the geometric mean of true positive rate (TPR) and true negative rate (TNR), where positive and negative corresponds to majority and minority classes respectively, to measure the performance. This measure is known as the g-mean and is calculated as (7).

$$\text{g-mean} = \sqrt{TPR \times TNR} \qquad (7)$$

G-mean is an appropriate measure for class imbalance problems as TPR and TNR are calculated separately for each class. Also, g-mean is a recommended measure among others for general evaluation in binary classification [37].

We compare the synthetic over-sampling methods and the base model using the average of g-mean for the best performing classifier. Then, we check the statistical significance of the differences for every combination of classifiers between SOMM and the best alternative method amongst other seven using the Bayesian signed test [38]. The Bayesian signed test enables the comparison of two methods over multiple data sets. It has been used in several studies to compare between the proposed method and alternatives [15], [39]. Before conducting the comparison, we need to find an appropriate optimal value for the only parameter of SOMM to be used in generating synthetic instances. We check three different values as 5, 10 and 15 for this parameter.

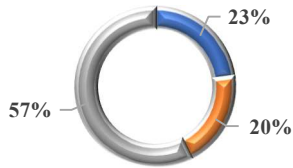

Fig. 6. The number of data sets on which each parameter performs the best in all three imbalance levels.

### B. Results

We obtain experimental results for 26 data sets over three different imbalance levels using seven classifiers with 30 repeats.

*1) Setting the parameter*

We need to define the number of nearest neighbours of the generated instance (k) described in step 4. We apply SOMM by setting k as 5, 10 and 15 on all 20 data sets and three different imbalance levels and conduct prediction using all seven classifiers. We compare these settings using g-mean based on the best performing classifier. The results show that k=15 outperforms the others for more data sets within each imbalance level. Figure 6 shows the success percentage of each setting in total for the three levels.

*2) Diversity comparison*

Table II presents the results of measuring the coverage diversity for over-sampling methods across 6 synthetic data sets (SDs) and three different numbers of minority instances. We report the mean of CovDiv measure and its standard deviation (SD) over 30 repeats (mean±SD). The greatest average value is indicated as bold, and if the value is statistically greater than other values, it is indicated as italic.

The results show that SOMM in 16 out of 18 combinations has the greatest value and in 12 of them is statistically greater than others. Only in two cases, SD4 and SD6 with 6 minority instances, SWIM can provide better diverse coverage than SOMM (not statistically significant). This shows SWIM can generate more diverse synthetic instances within the minority space when the minority space is convex and inside the majority space and the number of minority instances is low. Otherwise, SOMM provides superior diversity coverage within the minority space in different situation including non-convex minority space (Addressing RQs 1 & 2).

*3) Classification comparison*

We summarise the results by reporting the g-mean averages for the best performing classifier, i.e., the best g-mean value among seven classifiers, using synthetic over-sampling methods and the base model for both synthetic (SD1-6) and real (D1-20) data sets and each imbalance level in Tables III-V.

TABLE III
THE BEST G-MEAN VALUE OBTAINED BY SYNTHETIC OVER-SAMPLING METHOD FOR HIGH IMBALANCE LEVEL (5%)

| Dataset | Base | SOMM | SMOTE | MAHAKIL | SWIM | ADASYN | DBSMOTE | MWMOTE | DIWO |
|---|---|---|---|---|---|---|---|---|---|
| SD1 | 0.951 | 0.972 | 0.972 | **0.979** | 0.936 | 0.968 | 0.956 | 0.951 | 0.931 |
| SD2 | 0.903 | **0.984** | 0.973 | 0.980 | 0.947 | 0.981 | 0.972 | 0.979 | 0.978 |
| SD3 | 1 | 1 | 1 | 1 | 1 | 1 | 1 | 1 | 1 |
| SD4 | 0.939 | 0.981 | 0.981 | 0.979 | 0.977 | 0.978 | 0.974 | 0.978 | **0.982** |
| SD5 | 0.645 | **0.942** | 0.937 | 0.931 | 0.891 | 0.932 | 0.934 | 0.934 | 0.938 |
| SD6 | 0.622 | 0.935 | 0.923 | 0.935 | **0.939** | 0.935 | 0.918 | 0.919 | 0.935 |
| D1 | 0.963 | 0.964 | 0.964 | 0.964 | 0.964 | **0.967** | 0.96 | 0.965 | 0.964 |
| D2 | 0.786 | 0.843 | 0.84 | **0.844** | 0.843 | 0.817 | 0.798 | 0.836 | 0.782 |
| D3 | 0.842 | 0.974 | 0.97 | 0.972 | 0.961 | 0.964 | 0.953 | 0.967 | **0.982** |
| D4 | 0.563 | 0.945 | 0.944 | 0.952 | 0.911 | **0.958** | 0.928 | 0.949 | 0.935 |
| D5 | 0.557 | 0.952 | 0.943 | 0.948 | 0.909 | **0.958** | 0.934 | 0.953 | 0.944 |
| D6 | 0.467 | 0.909 | 0.941 | **0.949** | 0.888 | 0.947 | 0.905 | 0.940 | 0.858 |
| D7 | 0.285 | 0.712 | 0.706 | 0.703 | 0.697 | 0.721 | 0.673 | **0.735** | 0.699 |
| D8 | 1 | 1 | 1 | 1 | 0.999 | 1 | 1 | 1 | 1 |
| D9 | 0.71 | **0.798** | 0.712 | 0.735 | 0.655 | 0.741 | 0.657 | 0.723 | 0.354 |
| D10 | 0.752 | **0.949** | 0.924 | 0.916 | 0.745 | 0.937 | 0.913 | 0.939 | 0.915 |
| D11 | 0.881 | 0.838 | 0.943 | **0.945** | 0.929 | 0.934 | 0.886 | 0.901 | 0.861 |
| D12 | 0.437 | 0.617 | 0.659 | 0.703 | **0.715** | 0.701 | 0.68 | 0.692 | 0.635 |
| D13 | 0.612 | 0.631 | 0.698 | **0.731** | 0.62 | 0.72 | 0.706 | 0.725 | 0.697 |
| D14 | 0.663 | 0.744 | 0.796 | 0.807 | 0.658 | 0.807 | 0.793 | **0.811** | 0.752 |
| D15 | 0.429 | 0.604 | 0.649 | 0.702 | 0.593 | 0.685 | 0.663 | 0.633 | **0.704** |
| D16 | 0.68 | **0.772** | 0.699 | 0.685 | 0.68 | 0.726 | 0.708 | 0.693 | 0.753 |
| D17 | 0.207 | 0.404 | 0.415 | 0.418 | 0.422 | **0.505** | 0.453 | 0.393 | 0.434 |
| D18 | 0.949 | 0.95 | 0.947 | **0.978** | 0.977 | 0.958 | 0.941 | 0.949 | 0.965 |
| D19 | 0.708 | **0.921** | 0.833 | 0.872 | 0.735 | 0.872 | 0.841 | 0.872 | 0.845 |
| D20 | 0.569 | 0.589 | 0.63 | 0.689 | **0.69** | 0.687 | 0.666 | 0.662 | 0.649 |

TABLE IV
THE BEST G-MEAN VALUE OBTAINED BY SYNTHETIC OVER-SAMPLING METHOD FOR EXTREME IMBALANCE LEVEL (1%)

| Dataset | Base | SOMM | SMOTE | MAHAKIL | SWIM | ADASYN | DBSMOTE | MWMOTE | DIWO |
|---|---|---|---|---|---|---|---|---|---|
| SD1 | 0.696 | **0.941** | 0.902 | 0.896 | 0.914 | 0.839 | 0.842 | 0.854 | 0.896 |
| SD2 | 0.452 | **0.942** | 0.929 | 0.941 | 0.923 | 0.935 | 0.914 | 0.911 | 0.939 |
| SD3 | 0.997 | 1 | 0.997 | 0.999 | 1 | 0.999 | 0.999 | 0.999 | 1 |
| SD4 | 0.568 | 0.972 | **0.977** | 0.972 | 0.974 | 0.974 | 0.967 | 0.965 | 0.973 |
| SD5 | 0.114 | **0.915** | 0.886 | 0.894 | 0.872 | 0.904 | 0.895 | 0.780 | **0.915** |

| Dataset | Base | SOMM | SMOTE | MAHAKIL | SWIM | ADASYN | DBSMOTE | MWMOTE | DIWO |
|---|---|---|---|---|---|---|---|---|---|
| SD6 | 0.073 | 0.904 | 0.880 | 0.895 | **0.933** | 0.895 | 0.887 | 0.853 | 0.903 |
| D1 | 0.962 | **0.963** | 0.96 | 0.956 | **0.963** | 0.952 | 0.94 | 0.959 | **0.963** |
| D2 | 0.735 | **0.846** | 0.844 | 0.812 | 0.823 | 0.75 | 0.767 | 0.767 | 0.798 |
| D3 | 0.645 | **0.918** | 0.873 | 0.861 | 0.913 | 0.72 | 0.701 | 0.611 | **0.918** |
| D4 | 0.254 | **0.828** | 0.821 | 0.784 | 0.78 | 0.677 | 0.598 | 0.495 | 0.692 |
| D5 | 0.289 | **0.883** | 0.864 | 0.868 | 0.863 | 0.701 | 0.637 | 0.453 | 0.751 |
| D6 | 0.143 | **0.849** | 0.841 | 0.799 | 0.838 | 0.7 | 0.611 | 0.424 | 0.644 |
| D7 | 0.031 | 0.593 | 0.492 | 0.442 | **0.671** | 0.538 | 0.459 | 0.453 | 0.666 |
| D8 | 1 | 1 | 1 | 1 | 0.998 | 1 | 1 | 1 | 1 |
| D9 | 0.619 | **0.763** | 0.688 | 0.638 | 0.609 | 0.673 | 0.514 | 0.499 | 0.615 |
| D10 | 0.639 | **0.941** | 0.877 | 0.888 | 0.643 | 0.871 | 0.757 | 0.839 | 0.892 |
| D11 | 0.828 | 0.811 | 0.856 | 0.821 | **0.909** | 0.769 | 0.752 | 0.755 | 0.756 |
| D12 | 0.339 | 0.568 | 0.576 | 0.613 | **0.664** | 0.623 | 0.575 | 0.524 | 0.597 |
| D13 | 0.501 | 0.675 | 0.696 | **0.708** | 0.657 | 0.679 | 0.636 | 0.638 | 0.679 |
| D14 | 0.581 | 0.675 | 0.669 | 0.643 | 0.579 | 0.71 | 0.701 | 0.672 | **0.716** |
| D15 | 0.35 | 0.557 | 0.583 | 0.539 | 0.552 | **0.622** | 0.58 | 0.601 | 0.62 |
| D16 | 0.677 | **0.765** | 0.698 | 0.678 | 0.675 | 0.672 | 0.665 | 0.669 | 0.741 |
| D17 | 0.171 | 0.359 | 0.406 | 0.377 | 0.41 | **0.463** | 0.375 | 0.352 | 0.397 |
| D18 | 0.816 | 0.948 | 0.898 | 0.935 | **0.971** | 0.894 | 0.899 | 0.872 | 0.945 |
| D19 | 0.615 | **0.851** | 0.754 | 0.825 | 0.726 | 0.76 | 0.783 | 0.737 | 0.789 |
| D20 | 0.525 | 0.612 | 0.61 | 0.607 | **0.651** | 0.633 | 0.633 | 0.642 | 0.647 |

TABLE V
THE BEST G-MEAN VALUE OBTAINED BY SYNTHETIC OVER-SAMPLING METHOD FOR ABSOLUTE IMBALANCE LEVEL (SIZE 6)

| Dataset | Base | SOMM | SMOTE | MAHAKIL | SWIM | ADASYN | DBSMOTE | MWMOTE | DIWO |
|---|---|---|---|---|---|---|---|---|---|
| SD1 | 0.640 | **0.936** | 0.859 | 0.851 | 0.915 | 0.777 | 0.793 | 0.846 | 0.882 |
| SD2 | 0.341 | **0.939** | 0.920 | 0.936 | 0.915 | 0.926 | 0.876 | 0.846 | 0.930 |
| SD3 | 0.995 | **0.999** | 0.995 | 0.995 | **0.999** | 0.996 | 0.995 | 0.996 | 0.997 |
| SD4 | 0.410 | 0.968 | **0.971** | 0.968 | 0.965 | 0.969 | 0.954 | 0.953 | 0.969 |
| SD5 | 0.042 | **0.908** | 0.882 | 0.889 | 0.867 | 0.898 | 0.854 | 0.765 | **0.908** |
| SD6 | 0.024 | 0.891 | 0.874 | 0.881 | **0.929** | 0.886 | 0.878 | 0.833 | 0.892 |
| D1 | 0.957 | 0.953 | 0.953 | 0.951 | **0.962** | 0.952 | 0.94 | 0.959 | 0.96 |
| D2 | 0.725 | **0.764** | 0.722 | 0.638 | 0.751 | 0.757 | 0.77 | 0.679 | 0.728 |
| D3 | 0.635 | 0.749 | 0.684 | 0.649 | 0.842 | 0.694 | 0.666 | 0.627 | **0.883** |
| D4 | 0.235 | 0.664 | 0.651 | 0.603 | **0.712** | 0.646 | 0.571 | 0.437 | 0.52 |
| D5 | 0.202 | **0.709** | 0.663 | 0.596 | 0.685 | 0.672 | 0.617 | 0.396 | 0.634 |
| D6 | 0.156 | 0.676 | 0.672 | 0.618 | **0.722** | 0.676 | 0.584 | 0.400 | 0.632 |
| D7 | 0.037 | 0.538 | 0.421 | 0.382 | 0.635 | 0.45 | 0.361 | 0.287 | **0.642** |
| D8 | 1 | 1 | 1 | 1 | 0.998 | 1 | 1 | 1 | 1 |
| D9 | 0.553 | 0.612 | 0.6 | 0.507 | 0.603 | **0.669** | 0.51 | 0.429 | 0.606 |
| D10 | 0.416 | **0.813** | 0.647 | 0.544 | 0.716 | 0.641 | 0.531 | 0.636 | 0.721 |
| D11 | 0.783 | 0.78 | 0.781 | 0.764 | **0.888** | 0.769 | 0.752 | 0.755 | 0.726 |
| D12 | 0.337 | 0.59 | 0.579 | 0.557 | **0.644** | 0.623 | 0.574 | 0.604 | 0.574 |
| D13 | 0.145 | 0.529 | 0.513 | 0.499 | 0.52 | **0.559** | 0.503 | 0.437 | 0.521 |
| D14 | 0.583 | **0.677** | 0.666 | 0.644 | 0.584 | 0.674 | 0.664 | 0.629 | 0.595 |
| D15 | 0.297 | 0.504 | 0.539 | 0.489 | 0.507 | **0.578** | 0.515 | 0.487 | 0.558 |
| D16 | 0.675 | **0.762** | 0.694 | 0.658 | 0.67 | 0.672 | 0.665 | 0.669 | 0.733 |
| D17 | 0.171 | 0.363 | 0.407 | 0.372 | 0.411 | **0.463** | 0.375 | 0.361 | 0.354 |
| D18 | 0.46 | 0.788 | 0.91 | 0.841 | **0.932** | 0.918 | 0.886 | 0.919 | 0.901 |
| D19 | 0.791 | **0.878** | 0.708 | 0.876 | 0.735 | 0.876 | 0.799 | 0.702 | 0.765 |
| D20 | 0.548 | 0.591 | 0.588 | 0.547 | 0.634 | 0.652 | 0.649 | 0.559 | **0.709** |

Table III shows that SOMM performs the best for synthetic data sets as it achieves the highest values for three data sets. SOMM performs competitive for other SDs as well. By considering real data sets, MAHAKIL is superior in comparison with other methods by achieving the best performance in 6 data sets. After that, SOMM and ADASYN perform the second best with having the best g-mean value for 5 data sets. When we look at all 26 data sets, SOMM and MAHAKIL equally stand at the first rank for the high imbalance level (5%).

Table IV demonstrates that SOMM has the best performance amongst synthetic over-sampling methods for synthetic data sets as it has the highest values for four data sets. For real data sets, SOMM outperforms other methods by having superiority for 11 data sets. SWIM comes the second by having the best g-mean value in 6 data sets. By considering all 26 data sets, SOMM stands superior for the extreme imbalance level (1%)

Table V shows that SOMM performs the best for synthetic data sets as it achieves the highest values for four data sets. By considering real data sets, SOMM is superior by being the best for 7 data sets. The second best is SWIM by achieving the best g-mean value for 6 data sets. Also, by looking at all 26 data sets, SOMM stands out amongst other methods for the absolute imbalance level (6 minority instances).

The overall figures across all three imbalance levels demonstrate the superiority of SOMM in comparison with other methods as being the best for all three levels. The performance of MAHAKIL drops significantly from the high imbalance level to extreme and absolute levels. Also, SWIM's performance deteriorates in comparison with other methods by increasing the number of minority instances. Thus, SOMM shows a more consistent performance across all three imbalance levels (addressing RQ4).

*4) Statistical testing*

We also compare the performance of synthetic over-sampling methods using the Bayesian signed test. Figure 7 shows the posterior probabilities to examine how SOMM can perform in comparison with its alternatives. The question is whether the best alternative (ALT) can outperform SOMM significantly over all combination of classifiers and data sets for different imbalance levels. The results show that the posterior probabilities are located in the neutral (rope) area, which denotes SOMM performs competitively in comparison with state-of-the-art methods.

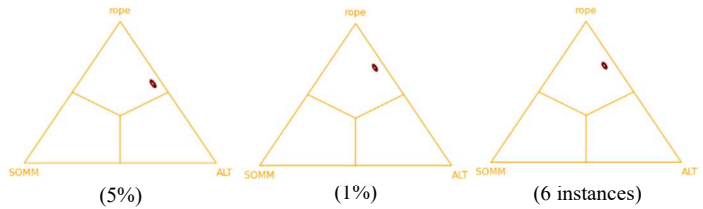

(5%)　　(1%)　　(6 instances)

Fig. 7.　Statistical test results.

*5) Discussion*

Considering synthetic data sets, SOMM performs the best when the majority class is bi-modal (SD2) and the minority space is non-convex (SD1 & SD5). However, SWIM is superior when minority space is convex and overlaps with the high-density region of the majority space (SD6).

By looking at results for real datasets in Tables III-V, we can see there are a few data sets for which SOMM does not perform best for any imbalance level in comparison with other methods. On the other hand, there are some data sets for which SOMM always performs best. In this section, we visualise two examples from each group using the first two PCs to understand why SOMM works well or otherwise. We visualise the instances of training and test subsets from both classes to show the data spaces, the training instances used for synthetic over-sampling and the real minority space (test subset) with which synthetic instances should correspond. We also generate synthetic instances using eight methods to provide an understanding on how these methods work.

TABLE VI
DATA SETS FOR MULTICLASS IMBALANCE PROBLEMS

| Dataset | Name | Class | Dimension | Size | Class description | Class distribution (%) | IR |
|---|---|---|---|---|---|---|---|
| MC-D1 | Wine Quality | 3 | 11 | 6,497 | 3,4/5,6,7/8,9 | 4/76/20 | 24.8 |
| MC-D2 | Wine Quality | 5 | 11 | 6,497 | 3,4/5/6/7/8,9 | 4/33/44/17/3 | 52.0 |
| MC-D3 | Wine Red Quality | 4 | 11 | 1,599 | 3,4/5/6/7,8 | 4/42.5/40/13.5 | 25.2 |
| MC-D4 | Wine White Quality | 5 | 11 | 4,898 | 3,4/5/6/7/8,9 | 3.5/30/45/18/3.5 | 49.8 |
| MC-D5 | Vehicle | 3 | 18 | 846 | Opel,Saab/Bus/Van | 51/26/23 | 2.3 |
| MC-D6 | Vowel 3 | 3 | 10 | 990 | 0/1,2,3,4,5,6,7,8,9/10 | 9/82/9 | 16.2 |
| MC-D7 | Vowel 5 | 5 | 10 | 990 | 0/1/2,3,4,5,6,7,8,9/10 | 9/9/64/9/9 | 24.4 |
| MC-D8 | Forest Cover | 6 | 54 | 3,000 | 1/2/3/4,5/6/7 | 34/50.5/6/2.5/3/4 | 93.7 |
| MC-D9 | Poker | 6 | 10 | 5,000 | 0/1/2/3/4/5,6,7,8,9 | 50/43/4.3/2/0.3/0.4 | 637.0 |
| MC-D10 | Abalone 9-18 | 3 | 9 | 2,676 | 9-13/14,15/16-18 | 85/9/6 | 22.1 |

Figure 8 and 9 (Figures 8-12 are provided as supplementary materials) show the plots for data sets D15 and D18, respectively, where always a synthetic over-sampling method other than SOMM performs best. In D15, the spaces of minority and majority classes significantly overlap, but there are few minority instances out of the overlapping space (outliers). As SOMM tries to generate synthetic instances outside of the majority neighbourhood, it generates many instances close to outliers when outliers are amongst the training's minority instances. In D18, the minority space is convex. Thus, all methods including SOMM can perform well to generate synthetic instances inside the minority space. That is the reason MAHAKIL is the superior method when the imbalance level is high. However, another characteristic for this data set is that the distributional information of the classes is helpful to separate two classes. So, SWIM can cover the minority space by generating instances on the hyper-elliptical density contours when number of minority instances in training data set decreases. This is consistent with what is observed in SD6.

Figures 10 and 11 show the plots for data sets D16 and D19, respectively, where SOMM is always superior. In D16, the spaces of minority and majority classes overlapping but not significantly. Also, the minority space is bigger than the convex hull formed by the minority instances in training subset. Thus, SMOTE family and MAHAKIL fail to generate instances to cover the real space of the minority class. MAHAKIL also generates less diverse instances even within the convex hull. On the other hand, SWIM widens the minority space mistakenly as the distributional information of majority class is not informative for the minority class. However, SOMM can generate synthetic instances diversely within the real space of the minority class. In D19, the minority space is convex, but it has a long tail. SOMM is the only method that can cover the real space of the minority class successfully. It is worth mentioning that SWIM fails to generate synthetic instances using the original data for several data sets (including D16 and D19) as the inverse matrix cannot be computed. Thus, we calculate the QR-decomposition of the majority subset to identify independent features. Then, SWIM is applied to the reduced space as instructed in [15].

We also show PC plots of D8 to provide further insights. Figure 12 shows that the two classes are completely separated and classifiers like SVM (with radial kernel) can easily distinguish between two classes even using the base model. Although ADASYN, DBSMOTE and MAHAKIL do not provide diverse coverage over the real minority class, the classification is successful. However, SWIM generates instances that are misleading for classifiers due to the tri-modal space of the majority class. It indicates that relying only on the distributional information of the majority class to generate synthetic instances is not sufficient. Figures 7-9 also show SOMM, SWIM and DIWO do not generate the synthetic instances using linear interpolations like SMOTE family.

## IV. EXPERIMENT RESULTS FOR MULTICLASS IMBALANCE PROBLEMS

### A. Experimental design

We design experiments to compare the performance of SOMM with MDO and SMOM methods for binary imbalance problems. We do not include SMOTE, MAHAKIL and SWIM as they are methods for binary imbalance problems. Using SMOTE and MAHAKIL for multiclass problems would increase the chance of over-generalisation. MDO and SMOM have been shown superior in comparison with the SMOTE family [31], [32]. The performance of SWIM decreases for multiclass imbalance problems because the probability of having multi-modal majority class (classes other than the minority class) increases. We set the parameters for MDO as $K1=10$ and $K2=5$ as suggested in [31], and for SMOM $k1=12$, $k2=8$, $rTh=5/8$, $nTh=10$, $w1=0.2$, $w2=0.5$, $r1=1/3$ and $r2=0.2$ as suggested in [32]. We set the parameter for SOMM as $k=15$ as it is shown in Fig. 5.

We use 10 benchmark data sets (listed in Table VI) in this experimental design. These data sets have various ranges of size, dimension, number of classes and Imbalance Ratio (IR), which is calculated as (8).

$$IR = \sum_{i=1}^{C-1} \sum_{j>i} \left(\frac{N_i}{N_j} - 1\right) \tag{8}$$

where $C$ is the number of classes, $N_i$ and $N_j$ refer to the number of instances for class $i$ and $j$ respectively, classes are sorted in a way that if $j > i$, $N_i >= N_j$.

We divide the data sets into training and testing subsets in the same way as for binary imbalanced data sets and repeat the experiments 30 times. We use the same classifiers except GLM, which is only used for binary classification, to construct classifiers on imbalanced training subset (base model) and balanced training subsets by SOMM, MDO and SMOM. We measure the performance of the classifiers on the testing subsets MG [40], which is a common measure for multiclass classification. MG is calculated as (9).

$$MG = (\prod_{i=1}^{C} Recall_i)^{\frac{1}{C}} \quad (9)$$

where $Recall$ for each class is the $TPR$ for that class ($\frac{TP}{TP+FN}$).

We compare the synthetic over-sampling methods and the base model using the average of MG for the best performing classifier. We also investigate the runtime complexity of SOMM, MDO and SMOM because we have outlined the lower computation as one of the advantages of SOMM over others.

*B. Results*

Table VII presents the average of MG for the best performing classifier of the base model, SOMM, MDO and SMOM for 10 data sets. The results show that SOMM is competitive for multiclass imbalance problems as it outperforms the base model and MDO for all data sets. SOMM cannot outperform SMOM for some data sets. However, SOMM has achieved higher values for three data sets, it has the same value for one data set, and it has achieved close values for three data sets.

TABLE VII
THE BEST MG VALUE OBTAINED BY SYNTHETIC OVER-SAMPLING METHODS FOR MULTICLASS IMBALANCE DATA

| Dataset | Base | SOMM | MDO | SMOM | Dataset | Base | SOMM | MDO | SMOM |
|---|---|---|---|---|---|---|---|---|---|
| MC-D1 | 0.439 | 0.477 | 0.423 | **0.547** | MC-D6 | 0.900 | 0.974 | 0.804 | **0.981** |
| MC-D2 | 0.415 | 0.480 | 0.419 | **0.513** | MC-D7 | 0.888 | 0.965 | 0.831 | **0.975** |
| MC-D3 | 0.396 | 0.490 | 0.371 | **0.492** | MC-D8 | 0.145 | **0.494** | 0.256 | 0.453 |
| MC-D4 | 0.453 | 0.518 | 0.454 | **0.525** | MC-D9 | 0 | **0.067** | 0.011 | 0.045 |
| MC-D5 | 0.961 | **0.969** | 0.795 | **0.969** | MC-D10 | 0.157 | **0.321** | 0.249 | 0.298 |

Table VIII presents the average runtime (in seconds) for the best performing classifier for SOMM, MDO and SMOM for 10 data sets. SOMM outperforms both MDO and SMOM by achieving the lowest runtime values in all data sets. SMOM has the highest runtime values in all data sets and is on average more than 200 times slower than SOMM.

TABLE VIII
THE AVERAGE OF RUNTIME IN SECONDS BY SYNTHETIC OVER-SAMPLING METHODS

| Dataset | SOMM | MDO | SMOM | Dataset | SOMM | MDO | SMOM |
|---|---|---|---|---|---|---|---|
| MC-D1 | **166** | 4,367 | 69,466 | MC-D6 | **16.7** | 58.4 | 995.1 |
| MC-D2 | **225** | 8,301 | 81,345 | MC-D7 | **29.6** | 115.1 | 2,818.5 |
| MC-D3 | **19.4** | 478 | 2,134 | MC-D8 | **510.4** | 3,961 | 97,789 |
| MC-D4 | **129** | 4,679 | 52,567 | MC-D9 | **1,140** | 7,834 | 125,762 |
| MC-D5 | **5.3** | 169 | 858.1 | MC-D10 | **184.5** | 342.9 | 64,901 |

Although SOMM cannot outperform SMOM in terms of the MG metric for all data sets, the runtime complexity of SOMM is much less than SMOM. For example, SOMM can improve the base model in 166 seconds on average while SMOM can do so in 69,466 seconds (about 19 hours) on average.

Overall, the experimental results validate the performance of SOMM for both binary and multiclass imbalance problems.

V. CONCLUSION

We have proposed a method to generate synthetic instances to alleviate the class imbalance problem in classification by over-sampling the minority class. Reviewing the limitations of existing synthetic over-sampling methods, we have defined four research questions (RQ1-4) for this study. The proposed method, SOMM, generates synthetic instances by considering the information of both majority and minority classes. So, it does not generate instances in the regions belonging the majority class. This is particularly useful when the minority data space is non-convex, or the majority data space is not unimodal. This has been tested using synthetic data sets and discussed by visualisations (addressing RQ1). SOMM generates synthetic instances diversely cover the minority space. This property is examined using synthetic data sets and a proposed measure (addressing RQ2). This property provides an opportunity for generating synthetic instances that correspond with the real data space of the minority class and consequently are useful for constructing robust classifiers. SOMM updates the generated instances adaptable to both majority and minority data spaces (addressing RQ3). This adaptability helps that the proposed method performs well for either binary or multiclass imbalance problems.

We have examined the performance of SOMM in comparison with existing methods for binary imbalance problems using 6 synthetic data sets and 20 real data sets with high (around 5%), extreme (around 1%) and absolute (few minority instances) imbalance levels. The results show that SOMM outperforms other methods for extreme imbalance data, while remaining competitive for high and absolute imbalance data (addressing RQ4).

The main strengths of SOMM are to perform effectively for both binary and multiclass imbalance problems as well as being effective for different imbalance levels even few minority instances. We have also provided a detailed discussion of the results where SOMM outperforms other methods using PC plots. We have observed that SOMM can effectively generate diverse synthetic instances within the minority data space for real data sets. SOMM also handles the situation where the majority class does not have a unimodal distribution.

We have investigated the results where SOMM does not stand out in comparison with other methods to identify the limitations of the proposed method. We have recognised the limitation of SOMM in generating synthetic instances heavily around the outliers when the spaces of minority and majority classes are significantly overlapped. We suggest overcoming this limitation as a future research by considering patterns and density of minority instances. SOMM does not perform outstanding in comparison with SWIM when distributional information is beneficial for generating synthetic instances. Another suggestion for further research is improving SOMM by considering distributional information.

SOMM is beneficial for researchers/practitioners to improve the performance of their classifiers for imbalanced data with various characteristics. It can perform effectively when the number of minority instances is low, while other re-sampling methods mostly fail to improve classifiers' performance. It can work properly for both binary and multiclass imbalance problems. Whereas other re-sampling methods are mostly effective for either binary or multiclass classification tasks. In case of multiclass imbalance problems, SOMM improves the classification with low time complexity.

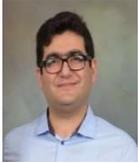

Hadi A. Khorshidi received his PhD from Monash University, where he worked as Research Fellow and Senior Data Analyst. He currently works as a Research Fellow in School of Computing and Information Systems at the University of Melbourne. His research areas include Decision Making and Optimization, Data Mining and Machine Learning, Modelling, Uncertainty and Digital Health, where he has published several papers.

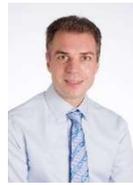

Uwe Aickelin holds a PhD degree from the University of Wales (UK). He is now a Professor and Head of School of Computing and Information Systems at the University of Melbourne. He is an Associate Editor of IEEE Transactions on Evolutionary Computation. His Research Interests include Artificial Intelligence (Modelling and Simulation), Data Mining and Machine Learning (Robustness and Uncertainty), Decision Support and Optimisation (Medicine and Digital Economy) and Health Informatics (Electronic Healthcare Records).